\documentclass[letterpaper, 10 pt, journal, twoside]{ieeetran}
\IEEEoverridecommandlockouts   
\usepackage{cite}
\usepackage{amsmath,amssymb,amsfonts}
\usepackage{algorithmic}
\usepackage{graphicx}
\usepackage{textcomp}
\usepackage{xcolor}
\usepackage{times}
\usepackage{hyperref}
\usepackage{dblfloatfix}
\usepackage{pifont}
\usepackage[margin=10pt,font=small,labelsep=space]{caption}
\def\BibTeX{{\rm B\kern-.05em{\sc i\kern-.025em b}\kern-.08em
    T\kern-.1667em\lower.7ex\hbox{E}\kern-.125emX}}
\begin{document}

\newcommand\blfootnote[1]{%
  \begingroup
  \renewcommand\thefootnote{}\footnote{#1}%
  \addtocounter{footnote}{-1}%
  \endgroup
}

\newcommand{\del}[1]{\textcolor{red}{#1}}
\newcommand{\add}[1]{\textcolor{blue}{#1}}
\newcommand{\rep}[1]{\textcolor{green}{#1}}
\markboth{IEEE Robotics and Automation Letters. Preprint Version. Accepted January, 2022}
{Can \MakeLowercase{\textit{et al.}}: Understanding Bird’s-Eye View of Road Semantics using an Onboard Camera} 

\title{\bf Understanding Bird’s-Eye View of Road Semantics using an Onboard Camera
}



\author{Yigit Baran Can$^{1}$ \thanks{$^{1}$Computer Vision Laboratory, ETH Zurich, Switzerland, {\tt\small\{yigit.can,alex.liniger,ozan.unal,paudel, vangool\}@vision.ee.ethz.ch} $^{2}$KU Leuven, Belgium} \quad Alexander Liniger$^{1}$\quad Ozan Unal$^{1}$ \quad Danda Paudel$^{1}$\quad  Luc~Van~Gool$^{1,2}$ }



\maketitle

\begin{abstract}
Autonomous navigation requires scene understanding of the action-space to move or anticipate events. For planner agents moving on the ground plane, such as autonomous vehicles, this translates to scene understanding in the bird's-eye view (BEV). However, the onboard cameras of autonomous cars are customarily mounted horizontally for a better view of the surrounding. In this work, we study scene understanding in the form of online estimation of semantic BEV maps using the video input from a single onboard camera. We study three key aspects of this task, image-level understanding, BEV level understanding, and the aggregation of temporal information. Based on these three pillars we propose a novel architecture that combines these three aspects. In our extensive experiments, we demonstrate that the considered aspects are complementary to each other for BEV understanding. Furthermore, the proposed architecture significantly surpasses the current state-of-the-art. Code: \url{https://github.com/ybarancan/BEV_feat_stitch}. 
\end{abstract}


\section{Introduction}

\IEEEPARstart{S}{cene} understanding on the image plane, using images and videos, is well studied with unparalleled success~\cite{cheng2020panoptic,xu2019spatiotemporal}. However, in many applications, this is not enough and the visual knowledge has to be mapped to the ``world" for understanding the scene. Some examples include 3D semantic modeling~\cite{riemenschneider2014learning}, augmented reality~\cite{ko2020novel}, and autonomous driving~\cite{jaritz20202d}. Such a mapping is particularly important when a mobile agent interacts with or navigates through the scene. In these cases, workspace understanding forms the basis of the decision making and navigation algorithms. This paper focuses on autonomous driving where the workspace is the ground plane where cars move. In fact, several works demonstrate the importance of the scene understanding directly on the ground plane in relation to the downstream tasks, such as motion prediction \cite{cui2019multimodal,zaech2020action,hong2019rules}, and planning \cite{DBLP:conf/rss/BansalKO19,chen2020learning}. 

\begin{figure}
    \centering
    \includegraphics[width=0.45\textwidth]{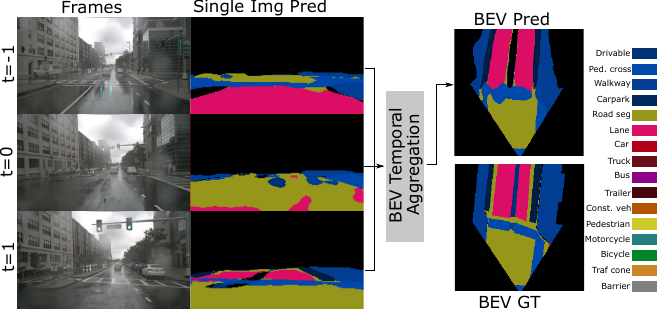}
    \caption{\small The proposed method takes video snippets and create BEV HD maps. It outputs an accurate segmentation of pedestrian crossing for the reference frame (t=0) even though pedestrian crossing is not visible in the reference frame.}
    \label{fig:teaser}
\vspace{-1.5em}
\end{figure}

Understanding the scene on the ground plane is fundamental, but autonomous vehicles acquire the visual information using one or more virtually horizontal onboard cameras, which is orthogonal to the ground plane. This discrepancy between image and ground plane creates a dichotomy of image and workspace understanding. One may argue that the understanding of the image can be mapped to the ground, given the homography between them. This work studies the effectiveness of such naive solutions and proposes better alternatives. More precisely, we focus on a complete Bird's-eye View (BEV) understanding by estimating semantic High Definition (HD) maps for autonomous driving \cite{nuscenes2019}, as well as dynamic objects such as cars, and pedestrians directly on the ground plane.

We study the case of a front-facing monocular camera for BEV ground plane understanding, focusing on HD-maps and objects. Representing HD-maps in the BEV is natural and commonly practiced~\cite{wu2020motionnet,paz2020probabilistic,yang2018hdnet,casas2018intentnet, hong2019rules, can2021structured, can2021topology}. On the other hand, understanding scenes using only a monocular camera is the holy grail of computer vision.
Up to our knowledge, there exists little to no work in the literature that proposes an end-to-end trainable method for online BEV understanding which leverages multiple input frames from one monocular camera. An overview of our setup is shown in Fig.~\ref{fig:teaser}. Besides the necessity of mapping the image information into the BEV, semantic BEV understanding comes with the additional challenge of multiple class labels per pixel. The multi-class labels are pivotal for the BEV and HD-maps since the same ground region may be used for different purposes. For example, a pedestrian crossing is also a drivable area. 

The traffic scene suffers from occlusion caused by dynamic agents as well as static obstacles such as buildings and trees. It, therefore, stands to reason that the temporal information in the video stream input should be exploited to eliminate occlusions as much as possible. While it is relatively straightforward to aggregate multiple frame outputs as post-processing to obtain semantic maps for static classes \cite{philion2020lift, DBLP:conf/cvpr/RoddickC20}, exploiting temporal information to boost segmentation of dynamic agents has never been tackled before.

\begin{figure}
    \centering
    \includegraphics[width=0.45\textwidth]{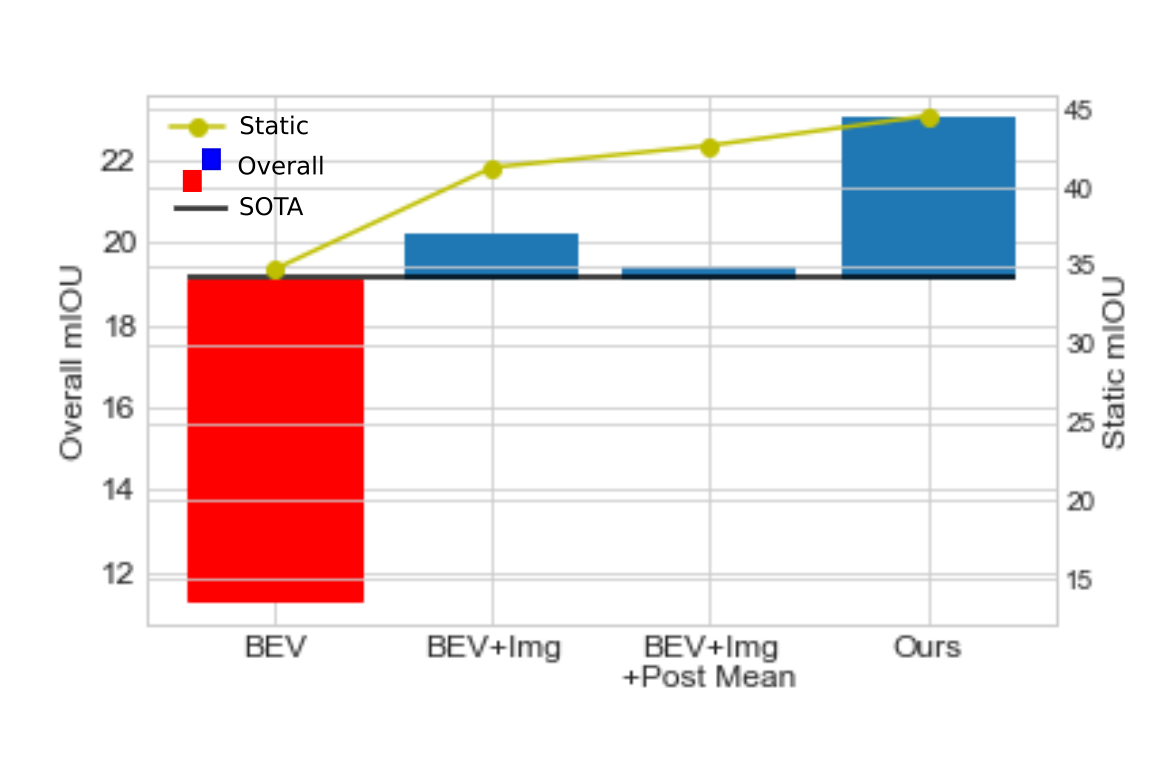}
    \vspace{-1.5em}
    \caption{An overview of the overall and static mIOU of the proposed components. Black line indicates SOTA~\cite{DBLP:conf/cvpr/RoddickC20} for both overall mIOU (left axis) and static mIOU (right axis). Post mean refers to taking mean over single frame estimates as a postprocessing step, see \ref{subsec:exp_abl}. Our method surpasses SOTA substantially in both overall and static only mIOU, even without temporal information. Proposed temporal aggregation and image branch boost performance drastically for object classes as well as static classes.}
    \label{fig:overviewPerformance}
    \vspace{-2em}
\end{figure}

We study three key aspects of processing the visual data for our task; image-level understanding, BEV understanding, and temporal data understanding. Such a study is made possible, thanks to our proposed deep neural network architecture. The proposed model simultaneously processes the visual information both in the image plane and the BEV, using BEV temporal aggregation, see Fig.~\ref{fig:teaser}. Our network can also exploit the temporal information in a coherent manner using the known camera motion. Our experimental results, an overview is reported in Fig.~\ref{fig:overviewPerformance}, show that the three aspects studied are complementary to each other. Our proposed method surpasses the state-of-the-art in both static and dynamic classes. Moreover, our proposed temporal aggregation module significantly boosts the performance of the inherently difficult dynamic object segmentation problem.

\noindent\textbf{Image-level understanding (Img):}
Since the observation takes place in the form of images, it is appealing to process the images as they are received. This is further bolstered by the advancements made on the side of image understanding, and the coherent spatial appearance in the image plane.

\noindent\textbf{BEV understanding (BEV):} Processing the visual data in the BEV is meaningful, as the understanding finally needs to be mapped in the same space. In essence, the output of the algorithm must be in BEV form. This is carried out by warping the image features and the processed image-level estimates onto the ground plane via a projective homography transformation, followed by BEV processing.  

\noindent\textbf{Temporal data understanding (Temp):}
The temporal data generally offers information for better scene understanding. This is particularly important for our task at hand because HD-maps are essentially defined on the static parts of the ground plane. Even though the ground plane can be occluded by dynamic objects, due to the dynamic nature of the ego camera and the objects, BEV temporal aggregation helps to remove such occlusions, see Fig.~\ref{fig:teaser}.

In summary, we propose a new method for BEV semantics understanding which fuses spatial-temporal information and operates both in the image and the BEV plane. The method is flexible and can work with any input video resolution and length as well as any BEV spatial output resolution. Our design allows us to study the importance of the three previously mentioned key aspects of BEV understanding. Thus, our major contributions can be summarized as follows. 
\begin{enumerate}
\setlength{\itemsep}{0pt}
\setlength{\parskip}{0pt}
\item We propose a novel deep neural architecture for BEV road semantics using monocular videos, for both static HD-map parts and dynamic objects understanding.  
\item We study the impact of independently and jointly learning in the image and/or the BEV plane, along with the temporal information. 
\item The results obtained by the proposed method are significantly superior to state-of-the-art methods.
\end{enumerate}

\section{Related Works}


Most existing methods for HD-maps perform visual understanding in the image plane~\cite{schulter2018learning,jang2018road}, followed by back-projection to BEV. Another category of methods implicitly exploits the knowledge of the ground plane, while performing the pixel-level understanding on the image plane~\cite{DBLP:journals/ral/LuMD19,DBLP:conf/cvpr/RoddickC20,reiher2020sim2real}. Some implicit methods also seek consistency across multiple synchronized cameras mounted on a rig~\cite{philion2020lift}.

Methods in the third category explicitly use the camera matrix to project intermediate features to ground plane ~\cite{DBLP:conf/cvpr/RoddickC20,philion2020lift}. Some other methods utilize additional information other than images such as LIDAR to produce multi-modal methods \cite{DBLP:journals/ral/PanSLAZ20,hendy2020fishing}. Unlike existing methods, we are interested in understanding the BEV directly using multiple frames of only one monocular camera.


The most relevant work for our paper is \cite{DBLP:conf/cvpr/RoddickC20}. In this paper, BEV HD-maps are generated from a single image. This is done by learning an implicit mapping from the image plane to the BEV plane by applying a dense transformer layer on the image features and resampling the processed feature map on the corrected BEV grid using the intrinsic camera matrix. This process is applied to a feature pyramid to combine the context information from higher levels with higher spatial resolution details from the lower levels of the backbone network. This approach is substantially different than ours since we can use multiple frames in our pipeline both during training and test using our BEV temporal aggregation module and we also exploit image-level understanding in a novel way to boost performance substantially. Moreover, our method can use any number of images at any resolution to output segmentation over a BEV region of any resolution and size while \cite{DBLP:conf/cvpr/RoddickC20} is constrained by the dense layers they employ.

Another similar work is MonoLayout \cite{DBLP:journals/corr/abs-2002-08394}, where the authors tackle BEV semantic segmentation from a single front camera image. They use a common encoder to extract features from the image and use separate static and object decoders to produce BEV estimations. To improve the visual results they utilize a GAN loss. However, they focus on very limited range of semantic classes (road+sidewalk and vehicle in KITTI and road and vehicle in Argoverse). Another focus is dealing with occluded regions, for which no labels exist in KITTI. Even though they also focus on BEV semantics, due to using small number of classes, and a different train/val split in Argoverse, it is hard to compare the results. 

\begin{figure*}[t]
\centering
\includegraphics[width=0.9\linewidth]{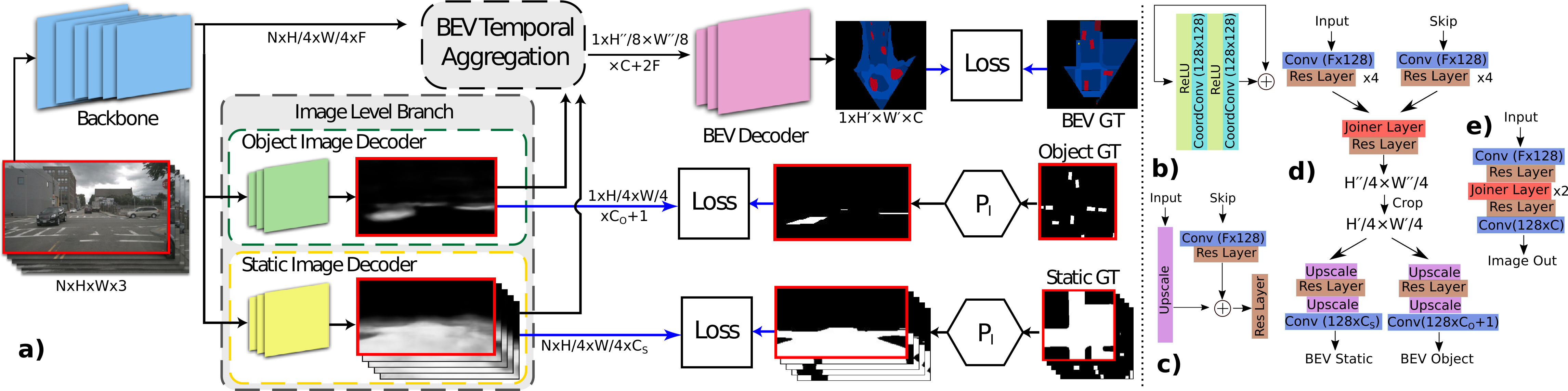}
\caption{ \small \textbf{a)} The overall method works with just one backbone. It takes N frames and the backbone features are obtained. The static image decoder produces segmentations for all frames while object estimations are only obtained for the reference frame. Temporal aggregation module takes static and object estimations as well as backbone features to output a combined feature map. BEV decoder processes this feature map to produce final estimates. Apart from BEV level, image level estimates are also directly supervised. Red borders outline reference frame related features and images. \textbf{b)} Residual layer, \textbf{c)} Joiner layer, \textbf{d)} BEV decoder, \textbf{e)} Image decoder.
}
\vspace{-1em}
\label{fig:shared_backbone}
\end{figure*}
Recently, Lift-Splat-Shoot \cite{philion2020lift} tackled the problem of generating semantic segmentation from a camera rig. However, they focused on estimating a few static HD-map classes (drivable area) and focus on dynamic objects. The key component of the method is estimating a discretized per-pixel depth distribution and implicitly warping the pixels by using this distribution.
Moreover, they use the true transformations to map the results from different cameras in the rig to a common BEV plane. It is possible to combine the object estimations from different camera images taken at the same time but it is not possible, at least trivially, to combine the object estimations from monocular images distributed in time due to object and ego-motion. 

\section{The Proposed Method}
The goal of our method is to predict the BEV semantic HD-map and dynamic objects for a specific reference frame, given an image sequence. As outlined in the introduction, we rely on three pillars, understanding the scene on the image plane, on the BEV plane, and finally exploiting temporal consistency. 
In our network the image level branch is responsible for the image plane understanding and the BEV Decoder and the BEV temporal aggregation modules for BEV and temporal reasoning.
An overview of the method is illustrated in Fig.~\ref{fig:shared_backbone}.

\noindent\textbf{Preliminaries.}
Our method accepts any number of frames with any resolution as input. Let us assume that the $N$ RGB input images are of dimensions $N\times H\times W \times 3$. The output of the network is a multi-channel BEV regular lattice grid which is of dimensions $1 \times H' \times W' \times C$. Where $H'$ and $W'$ are the spatial dimensions, and $C$ is the number of channels consisting of $C=C_S + C_O + 1$ with $C_S$ the number of static classes, $C_O$ the number of object classes plus an extra dimension for background. Since we aggregate multiple frames in a unified BEV grid, we need a BEV grid that spans the combined field of view (FOV), thus, we introduce an extended BEV grid which we denote with $''$ superscripts. 

\paragraph{Backbone.}
The input frames are fed to Cityscapes \cite{Cordts2016Cityscapes} pretrained Deeplabv3+ \cite{deeplabv3plus2018} backbone network to obtain feature maps of dimension $N\times H/8\times W/8\times 256$. We resize the feature maps to $N\times H/16\times W/16\times 256$ to increase the receptive field of the network and reduce memory demand. We keep the original feature map as a skip connection input. 

\subsection{Image Level Branch}

In the image level branch, the backbone features are processed by two decoders, the \textbf{static image decoder}, and the \textbf{object image decoder}, see Fig ~\ref{fig:shared_backbone}. Two branches are structurally similar and mainly differ in the final layer. Both decode the backbone features and generate a pixel-level segmentation, using skip connections from the backbone. More precisely the object branch outputs $C_O + 1$ object masks and the static branch $C_S$ static HD-map masks. The \textbf{object image decoder} learns the ground projections of objects and outputs estimates only for the reference frame, since we are only interested in the object locations in the reference frame while the \textbf{static image decoder} processes all the input frames. The output of the two decoders are heatmaps, which help guide BEV understanding. One can think of the two supervision in the image level branch as auxiliary task. These auxiliary supervisions are very helpful, as they offer intermediate guidance for both training and reasoning. 

\subsection{Temporal Aggregation}
\begin{figure}[h]
\centering
\includegraphics[width=0.95\linewidth]{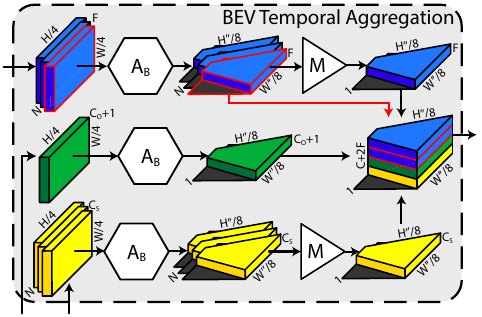}
\caption{ \small Temporal aggregation module combines information from all frames and all branches into one BEV feature map. Backbone features and image-level static estimates are projected with warping function $A_B$ to BEV and max (M) is applied in batch dimension. The results are concatenated in channel dimension. The reference frame backbone features (highlighted with red) are used in Max function as well as skip connection to concatenation.
}
\label{fig:temporal_agg}
\vspace{-2em}
\end{figure}

The temporal aggregation module is the core of our architecture and fuses the temporal information, which takes place directly on the BEV. This module consists of two key operations: (i) temporal warping and (ii) aggregation of the three inputs obtained from (1) the backbone, (2) the object decoder, and (3) the static map decoder. The aggregation process combines the image features from the backbone with the predictions from the object and static map decoders. Our aggregation process leads to the
BEV feature representation of size $1\times H''/8\times W''/8\times C + 2F$. An illustration of the BEV temporal aggregation module in shown in Fig.~\ref{fig:temporal_agg}.

\paragraph{Temporal warping} It is the process of warping all the image plane features and heatmaps into one common BEV representation, by means of a projective homography. This requires a per frame warping function $A_B$ that projects features from the $n$-th camera coordinate system to a fixed common BEV coordinate system. Such warping can be derived by assuming a flat ground plane.
Since our goal is to estimate the semantic BEV map of the reference frame, we use the ego-coordinate system of the reference frame as the fixed common BEV coordinate system.
The warp from the camera plane to the BEV plane is given as a homography between the two planes. Formally, for the $n$-th frame we define this using a warping matrix $W_n$, which transforms the image plane features or heatmaps $P_n$ to the common BEV map using the warping function $A_B(W_n,P_n)$. We can perform a warp since the camera intrinsic and extrinsics are known. Note that for the extrinsics, we assume that the motion of the car is known. In practice, the camera motion can be obtained using the vehicles' odometry pipeline, such as visual Simultaneous Localization and Mapping (vSLAM)~\cite{nister2004visual,durrant2006simultaneous,sualeh2019simultaneous,sumikura2019openvslam}. Given the extrinsic matrices at reference frame and frame $n$, $E_R$ and $E_n$ respectively, as well as the intrinsic matrix $K$ the warping function is $E_RE_n'K'$.

This process lets us generate BEV transformation of our image plane features and heatmaps, as shown in the middle part of Fig.~\ref{fig:temporal_agg}. Using the temporal warping function the backbone features are transformed into $N$ extended BEV girds with dimension $N\times H''/8\times W''/8\times F$, where $F$ is the number of features. Similarly, static heatmaps are transformed into $N$ extended BEV girds with dimension $N\times H''/8\times W''/8\times C_S$. For the object decoder where only one frame is processed, the estimates are mapped to a $1\times H''/8\times W''/8\times C_O+1$ grid. Note that, the extended BEV FOV enables the network to infuse information from the periphery into the target FOV.

\paragraph{Aggregation} To process the BEV with a common decoder we use \textbf{symmetric aggregation} to combine the temporally warped features. Since this aggregation should be independent of the order and number of input frames, we use symmetric functions, mainly max or mean aggregation. Thus, after performing the symmetric aggregation, we have three extended BEV grids: the object heatmap, the aggregated static heatmap, and the aggregated image features. To give the network more context about the current frame we also add the warped features of the current frame. Concatenating the four feature maps gives us our \textbf{temporal aggregated BEV feature maps} which we process with a BEV decoder to generate all BEV segmentations.  

Since warping of the image plane is only valid in the FOV of the camera, all BEV processing is done using \textbf{FOV masking}. Thus, after the features or heatmaps are warped to the BEV we mask them with the corresponding FOV mask. This also holds for our symmetric aggregation which is only performed if a pixel is in the FOV. More precisely let $B_n$ be the FOV mask of the $n$-th frame, then a max aggregation of the $N$ feature maps/heatmaps $P_n$ is defined as,
\begin{align}
    X_{ij} = \max\limits_t B_{t,ij} A(W_t, P_t)_{ij} \,,
    \label{eq:project}
\end{align}
where $X_{ij}$ is the max aggregated feature at the position $i,j$.

\subsection{BEV Decoder}
Given the temporal aggregated BEV feature map, the BEV decoder generates the final BEV predictions. Our decoder uses a residual network architecture to process the temporal aggregated BEV feature map. Besides standard residual blocks we also propose to use vertical residual blocks. Vertical residual blocks are similar to standard residual blocks, but the convolution kernels are set to $7 \times 1$. These blocks are used in between standard residual blocks to increase the FOV in the longitudinal direction (vertical matrix dimension). Since the spatial coordinates provide important clues for segmentation we utilized coordinate convolutions \cite{DBLP:conf/nips/LiuLMSFSY18} in all convolution blocks. This allows the network to better learn that roads are a highly structured spatial environment.

The assumption of flat ground plane can be problematic when the vehicle and part of the ground in the image are not on the same plane. However, using multiple frames, the probability that a given part of the ROI is on the same plane as the vehicle in at least one frame is higher. This allows the decoder to to use the correct frames for the corresponding regions and produce a coherent estimate.

\textbf{Warped skip connections} are obtained by applying the transformation given in Eqn \ref{eq:project} on the intermediate feature maps of the backbone to obtain BEV warped representation. The warped feature maps are used to upscale the processed feature maps to $1\times H''/4\times W''/4\times 128$. After the upscaling we crop the resulting extended FOV feature map to the FOV of the current frame resulting in feature maps of dimension $1\times H'/4\times W'/4\times 128$. The idea is that the heavy processing in the previous blocks infused information from the extended area into the current/target area. Thus, we can crop the feature map to the target area to keep memory consumption low. The target feature map is upscaled to the full size using bilinear upsampling and the static and object classes are estimated correspondingly.


\subsection{Loss and Occlusion Handling}
We supervise our network at three places, as visualized in Fig.~\ref{fig:shared_backbone}. First we supervise both image-level decoders directly using image-level labels. These labels are generated by the process explained in Subsection ~\ref{subsection:data_generation}. Briefly, we obtain image labels by warping BEV labels to the image plane. Therefore, the image pipeline learns the mapping between objects on the image plane and their projection on the BEV ground plane, see Fig.~\ref{fig:shared_backbone} for an illustration. This lets the method produce results without dealing with height/depth estimation of objects, as for example done in \cite{philion2020lift}. The main supervision is done in the BEV, where we directly supervise the semantic HD-map and the dynamic objects. All three losses are formulated as pixel-wise segmentations with focal loss \cite{DBLP:journals/pami/LinGGHD20} where $\gamma = 2$ and $\alpha$ is scaled according to the class frequency in the dataset.

For BEV supervision, occlusion handling is important, we already explained that we use FOV masks for the feature and heatmap warping. In the training, we apply loss on all visible region. However, we down-weight the loss in occluded regions. Since we use video, it is possible for our network to estimate occluded regions in the reference frame if these regions are visible at least in one frame. We estimate occluded regions using LiDAR following the protocol in \cite{DBLP:conf/cvpr/RoddickC20}. Following their definition, a BEV point is considered non-occluded if at least one LIDAR ray passes through it. However, we would like to highlight that no LIDAR data is used to generate our BEV estimates.

\section{Experiments}


We implemented our network in Tensorflow. During the training, we use $N = 4$ frames (one past and two future frames) and time interval between consecutive frames is randomly sampled between one and three frames. This lets the network better handle different speeds in motion and acts as a data augmentation. 

\subsection{Datasets}

In our experiments, we use the NuScenes \cite{nuscenes2019} and Argoverse \cite{DBLP:conf/cvpr/ChangLSSBHW0LRH19} datasets. NuScenes consists of 1000 sequences recorded in Boston and Singapore. The sequences are annotated at 2Hz and the dataset provides rich semantic HD-maps with static BEV segmentations such as drivable area, pedestrian crossing, and walkways. The dataset also provides 3D bounding boxes of 23 object classes. For our experiments, we select six HD-map classes and for dynamic objects, we follow \cite{DBLP:conf/cvpr/RoddickC20} and select 10 object classes. We only use the front camera for training and evaluation. 

We also conducted experiments on the Argoverse dataset which has 65 training and 24 validation sequences. In Argoverse the sequences are annotated at 10Hz. Unlike NuScenes, Argoverse only provides drivable area as a static HD-map class. For dynamic objects we again follow \cite{DBLP:conf/cvpr/RoddickC20}.


\begin{table}[h]
\vspace{-1em}
\tabcolsep=0.11cm
\scriptsize{
\begin{center}
\begin{tabular}{ |c|c|c|c|c|c|c|c|c|c| }
\rotatebox{50}{Model} & \rotatebox{50}{drivable} & \rotatebox{50}{ped cross} & \rotatebox{50}{walkway} & \rotatebox{50}{carpark}  & \rotatebox{50}{road} & \rotatebox{50}{lane} & \rotatebox{50}{occ} & \rotatebox{50}{4-Mean} & \rotatebox{50}{7-Mean} \\
\hline
PON & 60.4 & 28.0  & 31.0 & 18.4 & - & -& -& 34.5 & -\\ 
Deeplab & 58.9 & 14.3  & 19.1 & 21.2 & 55.0 & 40.4 & 48.0 & 28.4 & 36.7 \\
Ours & \textbf{72.9} & \textbf{31.2}  & \textbf{36.9} & \textbf{27.1} & \textbf{73.0} & \textbf{51.7} &
\textbf{61.1} & \textbf{42.0} & \textbf{50.6} \\ 
Ours(F) & 68.6 & 27.0  & 30.5 & 24.5 & 71.5 & 49.9 &
\textbf{61.1} & 37.6 & 47.6 \\ 
\hline
\end{tabular}
\end{center}
}
\vspace{-1em}
\caption{NuScenes static class only experiments are done in two settings. In the standard setting, results are evaluated only for non-occluded pixels. In the (F) FOV setting, every pixel in field-of-view is considered. Since our method processes videos, it can deal with occlusion. Our method surpasses the baselines in both settings, producing better results even when occluded pixels are considered.}
\vspace{-1em}
\label{tab:nuscenes_bev_results}
\end{table}


\subsection{Data Generation} \label{subsection:data_generation}
To provide a fair comparison with baselines, we followed the setup of Pyramid Occupancy Network \cite{DBLP:conf/cvpr/RoddickC20}. We used the same train/val split which is specially selected to provide a more challenging and representative sample distribution for the evaluation. We also used the same BEV lattice grid with a 0.25m resolution that spans a region of $[-25,25]$m in lateral and $[1,50]$m in longitudinal direction. Note that all our BEV labels are represented in this BEV grid. We also use the same occlusion masks to report the results.

Generating the lBEV labels for the static classes is straightforward since the datasets directly provide this data. To obtain the static image-level labels, we warp BEV labels onto the image plane. For the objects, we consider the ground plane projection of the 3D bounding boxes as the BEV labels of the object. Since we specifically want the image branch to estimate ground projections, we can obtain image-level labels by projecting these BEV labels to the image plane.

\subsection{Baselines}
\label{subsec:exp_abl}
To evaluate the performance of our method, we selected several baselines. The first baseline is a modified Deeplabv3+ \cite{deeplabv3plus2018} network that is pre-trained on Cityscapes \cite{Cordts2016Cityscapes}. We modified the final layers to match the number of classes and added a small sub-network for occlusion segmentation. The modified network is trained on NuScenes. The network outputs the image level segmentation, which we warp to the BEV plane using the ground truth projections. The second baseline is the SOTA method, Pyramid Occupancy Network (PON) \cite{DBLP:conf/cvpr/RoddickC20}. Finally, we also compare to Variational Encoder-Decoder Network (VED) \cite{DBLP:journals/ral/LuMD19} and View Parser Network (VPN) \cite{DBLP:journals/ral/PanSLAZ20} which generally perform worse than PON. As an additional baseline, we use our single frame network to obtain framewise BEV estimations. We then aggregate these estimations by warping them to the BEV reference frame and combining them by averaging, considering the individual visibility masks. This lets us benchmark our temporal aggregation module against the post-processing. We refer to this baseline as \emph{Post} in Tab.~\ref{tab:nuscenes_object_abla}.

\section{Results}


We evaluated our method against the state-of-the-art (SOTA) methods on two datasets, nuScenes \cite{nuscenes2019} and Argoverse \cite{DBLP:conf/cvpr/ChangLSSBHW0LRH19}. We look into two different settings, in the first, we focus on static HD-map classes, increasing the complexity compared to existing works by adding additional challenging classes. In the second setting we include dynamic objects, but for a better comparison use the same static HD-map classes as existing methods. The split is introduced to better show the benefit of our video-based method with challenging HD-map classes. 
Note that using video does not result in slow computation times since old information can be buffered, allowing our method to run with 21FPS on a RTX 2080Ti with an image size of ($448\times 800$).

\begin{figure}
    \centering
    \includegraphics[width=\linewidth]{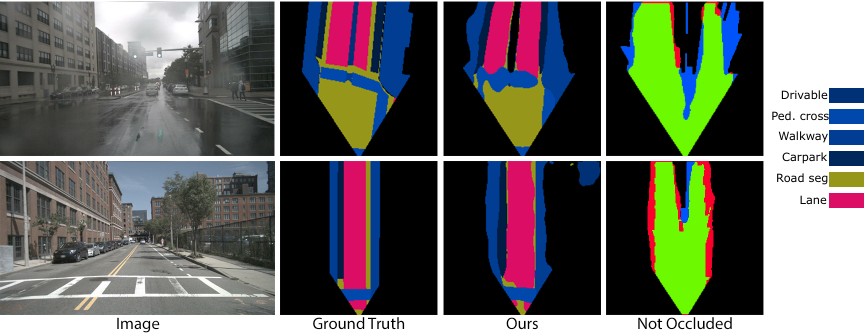}
    \vspace{-1em}
    \caption{Static examples with occlusion estimations. Green is true positive, blue is false negative and red is false positive.}
    \label{fig:our_nusc_static_results}
    \vspace{-1.5em}
\end{figure}


\subsection{Comparison against SOTA}

In Tab.~\ref{tab:nuscenes_bev_results}, we present the static only setting. This setting shows the power of the temporal component of the proposed method. Even when the occlusion mask is not used, marked with (F), the proposed method produces superior results compared to baselines. Since our method estimates two additional classes as well as the occlusion mask, we calculate the 4-Mean over the common classes for comparison as well as the 7-Mean over all the classes used in our network.

We present the results of our network in the static+objects setting in Tab.~\ref{tab:nuscenes_object_results}. It can be seen that the proposed method is superior in all static classes as well as several object classes, achieving a 4.5 mIOU improvement. An interesting observation is the performance increase in static classes in the static+object setting. When comparing Tab.~\ref{tab:nuscenes_bev_results} with Tab.~\ref{tab:nuscenes_object_results}, we can see that the results for the classes drivable and carpark area are improved. We assume that this is related to the link between these classes and the the dynamic classes. Since a region with a vehicle is likely a drivable area, and static cars are a clear indication for a carpark area. It is interesting that the additional supervision of dynamic objects resulted in such a clear improvement for these static classes through this indirect supervision signal. Note that all the shown nuScenes results use $N = 5$ frames (one past and three future) with an interval of three steps between the frames, see the ablation study in Section \ref{sec:ablation} for more details. 

\begin{table*}[h]
\scriptsize{
\begin{center}
\tabcolsep=0.11cm
\begin{tabular}{ |c|c|c|c|c|c|c|c|c|c|c|c|c|c|c|c|c|c| }
 \rotatebox{0}{Model} & \rotatebox{0}{Drivable} & \rotatebox{0}{Ped cross} & \rotatebox{0}{Walkway} & \rotatebox{0}{Carpark}  & \rotatebox{0}{Car} & \rotatebox{0}{Truck} & \rotatebox{0}{Bus} &\rotatebox{0}{Trailer}&\rotatebox{0}{Cons veh}&\rotatebox{0}{Pedest}& \rotatebox{0}{Motorcy}&\rotatebox{0}{Bicycle}&\rotatebox{0}{Traf cone}&\rotatebox{0}{Barrier}&\rotatebox{0}{Mean}\\
\hline
VED &  54.7 & 12.0  & 20.7 & 13.5 & 8.8 & 0.2 & 0.0 & 7.4&0.0 & 0.0&  0.0&0.0&0.0&4.0& 8.7 \\ 
VPN &  58.0 & 27.3  & 29.4 & 12.9 & 25.5 & 17.3 &
20.0 & 16.6 & 4.9&  7.1&5.6&4.4&4.6&10.8& 17.5 \\ 
PON &  60.4 & 28.0  & 31.0 & 18.4 & 24.7 & 16.8 &
20.8 & 16.6 & \textbf{12.3}&  \textbf{8.2}&7.0&\textbf{9.4}&\textbf{5.7}&8.1& 19.1 \\ 
Ours (Single frame) &  71.7 & 27.2  & 34.9 & 32.1 & 32.9 & 15.3 &
23.1 & 15.2 & 4.4 &  5.8 & 8.3 & 6.8 & 4.8 & \textbf{17.7} & 21.4 \\ 
Ours &    \textbf{74.9} & \textbf{31.6}  & \textbf{38.1} & \textbf{36.8} & \textbf{36.5} & \textbf{18.1} &
\textbf{27.4} & \textbf{17.6} & 4.4&  6.4&\textbf{8.6}&7.9&4.9& 17.2 &\textbf{23.6} \\ 
\hline
\end{tabular}
\end{center}
}
\vspace{-1em}
\caption{NuScenes Static+Object results. Ours (Single frame) is only the current frame and Ours is the full proposed method. Our method excels in static classes and surpasses state of the art in most object classes. }
\vspace{-1em}
\label{tab:nuscenes_object_results}
\end{table*}


Finally, in Tab.~\ref{tab:argoverse_object_results} we present our results for the static+object case using the Argoverse dataset. Similar to the other results we can see that our method excels in the static class, drivable area, where we improved the results drastically compared to the second best method. In the object classes, our method fails in small objects while performing on par or better in the large objects. 

\begin{table}[h]
\scriptsize{
\begin{center}
\tabcolsep=0.11cm
\begin{tabular}{ |c|c|c|c|c|c|c|c|c|c|c|c|c|c| }
 \rotatebox{0}{Model} & \rotatebox{0}{Driva} & \rotatebox{0}{Vehic} & \rotatebox{0}{Pedest} & \rotatebox{0}{Large}  & \rotatebox{0}{Bicyc} & \rotatebox{0}{Bus} & \rotatebox{0}{Trail} &\rotatebox{0}{Motor}&\rotatebox{0}{Mean}\\
\hline
VED &  62.9 & 14.0  & 1.0 & 3.9 & 0.0 & 12.3 & 1.3 & 0.0& 11.9 \\ 
VPN &  64.9 & 23.9  & 6.2 & 9.7 & 0.9 & 3.0 &
0.4 & 1.9 & 13.9\\ 
PON &  65.4 & \textbf{31.4}  & \textbf{7.4} & 11.1 & \textbf{3.6} & 11.0 &0.7 & \textbf{5.7} & 17.0 \\ 
Ours &    \textbf{79.8} & 28.2  & 4.8 & \textbf{11.4} & 0.0 & \textbf{22.5} &\textbf{1.1}  & 0.4 & \textbf{18.5}\\ 
\hline
\end{tabular}
\end{center}
\tabcolsep=0.11cm
}
\vspace{-1em}
\caption{Argoverse Static+Object results }
\vspace{-2em}
\label{tab:argoverse_object_results}
\end{table}








\subsection{Visual results}
\label{sec:vis_results}

Some examples for the static only network are given in Fig.~\ref{fig:our_nusc_static_results}. We estimate six static classes and occlusion. The results show that our method produces accurate and consistent estimations. 

\begin{figure*}[h]
    \centering
    
    \includegraphics[width=.8\textwidth]{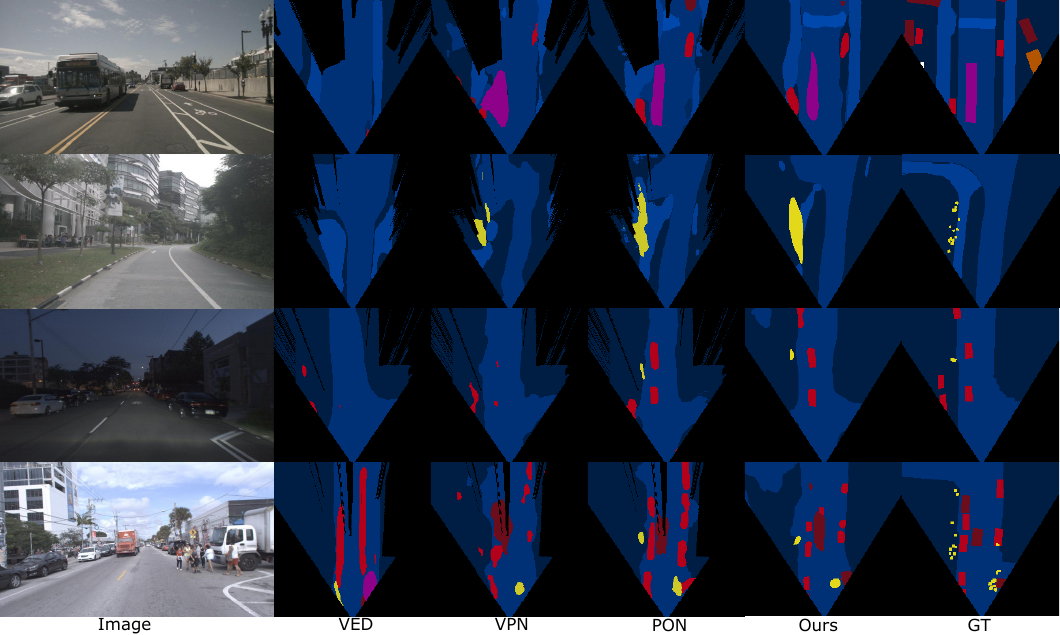}
    \caption{Comparison with SOTA on Nuscenes (upper 2) and Argoverser (lower 2). Competitors' results are adopted from \cite{DBLP:conf/cvpr/RoddickC20}. In all examples, our method capture static scene much more accurately. In the last row, our method can detect truck in the lower right, while PON classifies it as a car. On the upmost image, PON hallucinates a truck on the up right. See Fig.~\ref{fig:teaser} for the legend.}
    \label{fig:comparison}
    \vspace{-.5em}
\end{figure*}

Visual comparison against SOTA methods is given in Fig.~\ref{fig:comparison}, note that we used the same images as shown in \cite{DBLP:conf/cvpr/RoddickC20}. For all our visual results we do not apply the estimated occlusion mask, the same holds for the ground truth. But we apply the occlusion masks to the competitors' results. This presents a more complete picture of our predictions which are less limited by the FOV compared to the SOTA methods. In summary, static classes are more accurate and consistent while the object results are comparable or better. 

\begin{figure}[h]
\centering
\vspace{-0.5em}
\includegraphics[width=.8\linewidth]{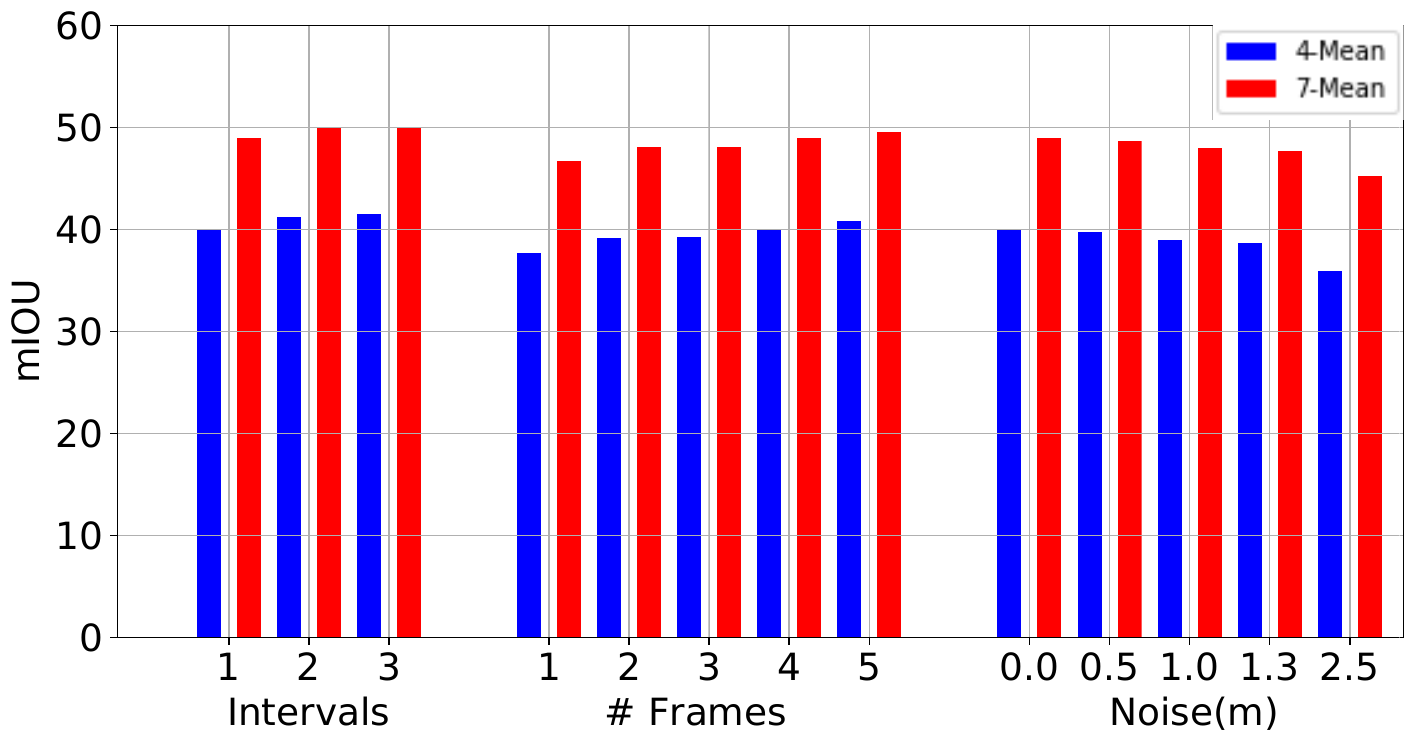}
\vspace{-0.5em}
\caption{ \small Frame intervals, number of frames and standard deviation (meters) of zero-mean Gaussian noise on the corresponding point locations that are used to calculate the projective transformations.}
\label{fig:ablation}
\vspace{-1.5em}
\end{figure}

\begin{table*}[!h]
\scriptsize{
\begin{center}
\vspace{0.5em}
\tabcolsep=0.11cm
\begin{tabular}{ |c|c|c|c|c|c|c|c|c|c|c|c|c| }
\hline
\multicolumn{4}{|c|}{Component} & \multicolumn{7}{c|}{IoU per class [\%]} & \multicolumn{2}{c|}{mIoU [\%]} \\
\hline
Image & BEV & Temp & Backbone & drivable & crossing & walkway & carpark  & road segment & lane & occlusion & 4-Mean & 7-Mean \\
\hline
\ding{51} & & &Deeplab & 58.9 & 14.3  & 19.1 & 21.2 & 55.0 & 40.4 & 48.0 & 28.4 & 36.7 \\
& \ding{51} & &Deeplab& 68.5 & 22.8  & 32.0 &24.7 & 66.8 & 47.2 & 58.5 & 37.0 & 45.8 \\ 
\ding{51} & &\ding{51} &Deeplab & 64.5 & 15.6  & 20.9 &24.4 & 61.6 & 45.6 & 51.6 & 31.4 & 40.6 \\
\ding{51} & \ding{51} & &Deeplab &  69.1 & 25.8  & 32.5 & 23.4 & 68.8 & 46.9 & 60.2 & 37.7 & 46.7\\ 
& \ding{51} & \ding{51} &Deeplab & 70.2 & 25.0  & 34.4 & 27.2 & 68.9 & 49.4 & 59.8 & 39.2 & 47.8 \\
\ding{51}& \ding{51} & \ding{51} &Deeplab & \textbf{71.3} & \textbf{28.2}  & \textbf{35.3} & {25.8} & \textbf{71.3} & \textbf{49.5} & \textbf{61.2} & 40.1 & \textbf{49.0} \\ 
\ding{51}& \ding{51} & \ding{51} &ResNet &  70.4 & 27.7  & 33.5 & \textbf{30.4} & 68.6 & 45.5 & 56.7& \textbf{40.5} & 47.6\\ 
\hline
\end{tabular}
\end{center}
}
\vspace{-1em}
\caption{Ablation study on the three different components of the proposed method shows that every component provides improvement. The biggest effect is observed when BEV processing is introduced. Temporal information provides 2 points boost and using the image-level branch on top of these result improves the best model with a further 1.2 points boost.}
\label{tab:main_ablation}
\end{table*}

\subsection{Ablation Studies}
\label{sec:ablation}

To measure the effect of the different components of the proposed method, we studied different settings of the static classes only network in Tab.~\ref{tab:main_ablation} and the static+object network in Tab.~\ref{tab:nuscenes_object_abla} on NuScenes. For all these results we use $N=5$ frames with a spacing of three frames. Note that the we ablate these settings afterwards.  


In our ablations, \emph{BEV} refers to existence of a dedicated BEV decoder that processes warped features/estimates. \emph{Image} refers to existence of image branch which produces image level estimates that are used by the BEV decoder, if a BEV decoder exists. \emph{Temp} refers to using our temporal module to combine information from multiple frames. This includes backbone features and image level estimations, if they exist. If \emph{Temp} is not present, only the reference frame is used. Note that in this case, there is no need to know the ego pose or transformations between frames.

In both settings each component provides a performance boost, and the performance with a Resnet50 is similar to the DeepLabV3+ encoder we use. This shows that performance gains compared to the SOTA mainly come from the architecture and not the backbone. When we investigate the separate components we can take the following conclusions.

\noindent\textbf{BEV:} Only using image level reasoning is clearly not sufficient for the task and the BEV warping and decoder are fundamental for our method.

\noindent\textbf{Image Branch:}
Adding image level supervision gives slight gains in the static setting but is crucial for dynamic objects increasing the mIOU by 8.9 points. The image branch acts as an anchor for the temporally aggregated features. Moreover, in both the static and static+objects cases, the image branch eases training of the network significantly. 

\noindent\textbf{Temporal Aggregation:}
Temporal aggregation gives a solid and consistent boost in all the static classes. However, in the case also dynamic objects are considered, temporal aggregation does not help without the image branch (see BEV+Temp vs BEV+Img+Temp in Tab.~\ref{tab:nuscenes_object_abla}). To further highlight that the in-network temporal aggregation is preferable, we compare our temporal aggregation module with the temporal post processing approach discussed in Subsection \ref{subsec:exp_abl}. Even though the post processing slightly improves the results for static classes, it is not as effective as our approach. For the dynamic classes it even reduces the performance whereas our in-network aggregation further improves the IOU.

In Fig.~\ref{fig:ablation}, we show further ablation results for the static only case, studying details related to the temporal component of our network. The study shows that spacing the frames further apart and using more frames is beneficial. Moreover, to test that we are not too dependent on our assumption of known frame positions we tested how well our network performs if we corrupted the positions with zero-mean iid Gaussion noise. While adding noise with a higher standard deviation decreases the performance, for realistic small noise levels the performance is not noticeably affected. 

In summary the ablation shows that all the components work in harmony to achieve our SOTA results. The ablation also shows the flexibility of our method, since the same network trained with $N=4$ frames can achieve strong results with one up to five frames, in fact in all these setting producing state-of-the-art BEV segmentations.



\begin{table*}[!h]
\scriptsize{
\begin{center}
\tabcolsep=0.10cm
\begin{tabular}{ |c|c|c|c|c|c|c|c|c|c|c|c|c|c|c|c|c|c|c|c|c| }
\hline
\multicolumn{4}{|c|}{Component} & \multicolumn{14}{c|}{IoU per class [\%]} & mIoU [\%] \\
\hline
BEV & Image & Temp & Post  & \rotatebox{0}{Drivab} & \rotatebox{0}{Ped cross} & \rotatebox{0}{Walkway} & \rotatebox{0}{Carpark}  & \rotatebox{0}{Car} & \rotatebox{0}{Truck} & \rotatebox{0}{Bus} &\rotatebox{0}{Trailer}&\rotatebox{0}{Cons veh}&\rotatebox{0}{Pedest}& \rotatebox{0}{Motorcy}&\rotatebox{0}{Bicycle}&\rotatebox{0}{Traf cone}&\rotatebox{0}{Barrier}&\rotatebox{0}{Mean} \\
\hline
\ding{51} &  & & &  67.7 & 19.3  & 27.1 & 25.0 & 5.7 & 3.8 & 4.7 & 2.6 & 0.0 & 0.3&  0.3&0.0&0.2&1.0& 11.3 \\ 
\ding{51}&  & \ding{51}&  &  70.7 & 22.7  & 30.9 & 30.0 & 5.9 & 4.3 & 5.5 & 2.6&0.0 & 0.3&  0.3&0.0&0.2&1.2& 12.5 \\ 
\ding{51}& \ding{51} & &  & 71.7 & 27.2  & 34.9 & 32.1 & 32.9 & 15.3 &
23.1 & 15.2 & \textbf{4.4} &  5.8 & 8.3 & 6.8 & 4.8 & \textbf{17.7} & 21.4 \\ 
\ding{51}& \ding{51} & & \ding{51}& 73.9 & 29.2 &  35.8 & 31.9 & 32.2 & 12.1 & 14.8 & 7.1 & 1.3 & 3.6 & 5.0& 2.3&
 4.7 & 17.3 & 19.4\\
\ding{51}& \ding{51} & \ding{51}&   &  \textbf{74.9} & \textbf{31.6}  & \textbf{38.1} & \textbf{36.8} & \textbf{36.5} & \textbf{18.1} &
\textbf{27.4} & \textbf{17.6} & \textbf{4.4}&  \textbf{6.4}&\textbf{8.6}&\textbf{7.9}&\textbf{4.9}&17.2&\textbf{23.6} \\ 

\hline
\end{tabular}
\end{center}
}
\vspace{-1em}
\caption{NuScenes Static+Object component ablation. BEV refers to BEV processing, Image refers to image level estimation branch and Temp refers to our in-network temporal aggregation module. Post refers to taking mean over the single frame estimates as a post-processing step, see \ref{subsec:exp_abl}.}
\vspace{-1.5em}
\label{tab:nuscenes_object_abla}
\end{table*}

\section{Conclusion}
In conclusion, we propose a novel architecture for BEV understanding of road semantics using video input from a front facing monocular camera, mounted on a vehicle. The network exploits three key aspects for BEV map estimation, understanding in the image plane, which allows to leverage mature algorithms in a spatial coherent image. Temporal data understanding, which allows us to better handle occlusions, a crucial point for HD-map understanding. Finally, BEV understanding, which allows us to aggregate all the temporal and image-plane information in a consistent fashion.
The understanding is then performed directly on the action space of the autonomous vehicle.
We show that our method drastically improves the SOTA results for static HD-map segmentation, while also helping for dynamic objects such as cars.

{\small
\bibliographystyle{ieee_fullname}
\bibliography{egbib}
}

\end{document}